\documentclass{article} 
\usepackage{iclr2027_conference,times}
\usepackage{float}

\usepackage{amsmath,amsfonts,bm}

\def\eqref#1{equation~\ref{#1}}

\def\1{\bm{1}}

\DeclareMathAlphabet{\mathsfit}{\encodingdefault}{\sfdefault}{m}{sl}
\SetMathAlphabet{\mathsfit}{bold}{\encodingdefault}{\sfdefault}{bx}{n}

\usepackage{hyperref}
\usepackage{url}

\usepackage{tikz}
\usetikzlibrary{positioning,fit,backgrounds,arrows.meta,calc}
\usepackage{subcaption}

\title{MINE: Entropy-coded Memory-efficient Inference for Large Language Models}

\author{Emil Sch\"atzle \& Markus P\"uschel \\
Department of Computer Science\\
ETH Zurich\\
Zurich, Switzerland \\
\texttt{emil.schaetzle@inf.ethz.ch, pueschel@inf.ethz.ch}
}

\iclrfinalcopy
\begin{document}

\maketitle
\lhead{Preprint}  

\begin{abstract}
We present MINE (Memory-efficient INference Engine), a lossless entropy coding approach that accelerates single-stream inference of Large Language Models (LLMs), where existing lossless compression methods yield speedups only for specific data types such as \texttt{BF16} and under additional assumptions. Single-stream decoding is strongly memory-bound, since every weight must be streamed from memory once per generated token and requests cannot be batched in local or low-latency deployments. At the core of MINE is dtANS-LLM, a GPU entropy coding scheme adapted from the prior dtANS to the needs of LLM inference. It exploits the non-uniform distribution of weights, which persists even after quantization, and exposes hyper-parameters that trade compression ratio against decoding cost. We auto-tune these parameters for each tensor and GPU to obtain the Pareto front between model size and decoding latency. MINE integrates into vLLM as a plugin and selects the best kernel configuration for a given memory budget. On Llama~3.1~8B~Instruct with a 2.1-bit EntQuant checkpoint, MINE achieves speedups of 2.52$\times$ and 1.48$\times$ over the fastest baseline on an RTX~4060~Ti and an RTX~5090, at effective bitrates of 2.24 and 3.43~bits per weight. MINE is available as open source.
\end{abstract}

\section{Introduction}
\label{sec:introduction}
The dramatically increasing use of Large Language Models (LLMs) makes highly efficient inference a crucial requirement. Inference requires streaming and decoding the entire LLM network from RAM while each weight is used only once per request, which is a highly memory-bound setting \citep{MemoryWall24}. In data center deployments this is mitigated by collecting many requests into batches, but other settings such as local LLMs or low-latency applications require good single-stream performance.
But even in general, the increasing size of LLMs and the current RAM scarcity pose the need for memory-efficient LLM representations.

The most straightforward approach is lower-precision floating-point representations to reduce the memory footprint of the weights while losing accuracy \citep{FP8_22,MXFP4Eval23,NVFP4_25}. Many post-training quantization methods can be seen as a continuation of this idea to custom grids \citep{RTN23} and non-trivial choices of levels \citep{GPTQ23,AutoRound24,AutoRound25,HQQ23}. Others adapt to the importance of the individual weights \citep{QTIP24,GGUF23,AWQ24}. An alternative is Quantization-Aware Training \citep{QAT18,ParetoQ25}, which can go as far as only supporting three states for each weight \citep{Ternary24}.

An orthogonal approach, which can also be combined with quantization, is lossless compression of LLMs. The weights of LLMs are not distributed uniformly, a property that also holds after applying most quantization methods \citep{ShannonBound26}. Lossless coding methods exploit these statistical redundancies \citep{NeuZip24,DFloat11_25,EntQuant26} to compress the weights, where the limit is given by the weights' entropy. They can increase inference throughput since the saved memory allows for larger batch sizes \citep{ShannonBound26} or using simplified coding schemes that leverage additional assumptions on the domain \citep{ZipServ26}. However, to date, none of these works provides a speedup using generic entropy coding for single-stream decoding on GPU. 

\paragraph{Contributions.} 
In this paper, we present MINE (Memory-efficient INference Engine), which uses fast lossless entropy coding to compress (quantized) LLMs and achieve inference speedup in the single-stream setting. We adapt an existing method \citep{dtANS26} of entropy coding for GPUs to the particular needs of inference and introduce hyper-parameters to adapt to different settings. Using auto-tuning we then compute the Pareto front between size and single-stream decoding latency to obtain the best possible latency for a given memory budget. We provide highly optimized GPU kernels: \texttt{mvm} to perform the needed matrix-vector multiplication and \texttt{dequant} to materialize tensors to global memory for prefill. Our specific contributions are as follows:
\begin{itemize}
    \item We present the entropy coding variant dtANS-LLM, adapted from the existing dtANS \citep{dtANS26} for the specific needs of single-stream inference in LLMs. 
    \item We auto-tune the hyper-parameters of dtANS-LLM for best performance given the characteristics of the tensors and the compute-to-memory-bandwidth ratios of the GPU.
    \item We integrate dtANS-LLM into the vLLM inference engine \citep{vLLM23} as a plugin which we name MINE. Given a memory budget, MINE selects the most promising combination of kernels for performance.
    \item We evaluate MINE on Llama~3.1~8B~Instruct \citep{Llama3_24}. On a low-entropy checkpoint (EntQuant \citep{EntQuant26} at 2.1~bits) it achieves speedups of 2.52$\times$ and 1.48$\times$ over the fastest baseline on an RTX~4060~Ti and RTX~5090, respectively. The effective bitrates are 2.24 and 3.43~bits per weight, showing that the optimal parameter choices indeed depend on the card.
    \item We make MINE available as an open-source repository.
\end{itemize} 

\section{Overview}

\begin{figure}[t]
\centering
\begin{tikzpicture}[
  font=\sffamily\scriptsize,
  >={Straight Barb[length=2.5pt,width=3.5pt]},
  every path/.style={line width=0.5pt},
  every node/.append style={execute at begin node={\hyphenpenalty=10000\relax}},
  io/.style={align=center, inner sep=1pt},
  step/.style={fill=white, text width=#1, align=center, inner sep=2.5pt, rounded corners=2pt},
  sub/.style={fill=black!12, text width=#1, align=center, inner sep=2.5pt, rounded corners=2pt},
  panel/.style={fill=blue!40!black, inner sep=4pt, rounded corners=2pt},
  title/.style={font=\sffamily\scriptsize\bfseries, inner sep=1pt},
  ptitle/.style={font=\sffamily\scriptsize\bfseries, text=white, inner sep=1pt},
  node distance=5pt and 0pt,
]
\def\pw{2.30in}   
\def\sw{2.217in}  
\def\gap{2.72in}  
\def\isep{3pt}    

\node[ptitle] (ttitle) at (0,0) {Tuning (once per checkpoint and GPU)};
\node[step=\pw, below=4pt of ttitle] (t1)
  {Enumerate codings per launch: weights per symbol (1/2/4),
   slice width $s_c$, table size $M$, \texttt{FP8} payload layout};
\node[step=\pw, below=of t1] (t2)
  {Prune by predicted compressed size};
\node[step=\pw, below=of t2] (t3)
  {Tune kept set $t$; encode with dtANS-LLM; tune warps per block for cold-cache \texttt{mvm} latency};
\node[step=\pw, below=of t3] (t4)
  {Tune \texttt{dequant} / \texttt{encode} / \texttt{decode} on speed--size Pareto front};

\node[ptitle] (stitle) at (\gap,0) {Serving (MINE plugin in vLLM)};
\node[step=\pw, below=4pt of stitle] (s1)
  {Pick fastest codings under the budget (Pareto front)};
\node[step=\pw, below=of s1] (s2)
  {Load and encode tensors one by one};
\node[title, below=5pt+\isep of s2] (s3title) {Per forward pass, per launch};
\node[sub=\sw, below=3pt of s3title] (s3a)
  {Batch 1: \texttt{mvm} fuses dtANS-LLM decoding with the matrix-vector product};
\node[sub=\sw, below=of s3a] (s3b)
  {Batch ${>}1$: \texttt{dequant} materializes the layer, then GEMM};

\path let \p1=(t4.south), \p2=(s3b.south) in
  coordinate (bot) at (0,{min(\y1,\y2-\isep)});

\begin{scope}[on background layer]
  \node[panel, fit=(ttitle)(t1)(t4)(ttitle|-bot)] (tpanel) {};
  \node[panel, fit=(stitle)(s1)(s3b)(stitle|-bot)] (spanel) {};
  \node[step=\pw, inner sep=\isep, fit=(s3title)(s3a)(s3b)] (s3) {};
\end{scope}

\node[io, anchor=south] (tin1) at ([xshift=-0.45in]tpanel.north) [yshift=11pt]
  {Checkpoint (\texttt{BF16} / \texttt{FP8} / quantized)};
\node[io, anchor=south] (tin2) at ([xshift=0.85in]tpanel.north) [yshift=11pt] {GPU};
\node[io, below=11pt of tpanel] (tout)
  {Tune file: size and latency\\ of every coding per launch};
\draw[->] (tin1) -- (tin1|-tpanel.north);
\draw[->] (tin2) -- (tin2|-tpanel.north);
\draw[->] (t1) -- (t2);
\draw[->] (t2) -- (t3);
\draw[->] (t3) -- (t4);
\draw[->] (tpanel) -- (tout);

\node[io, anchor=south] (sin1) at ([xshift=-0.85in]spanel.north) [yshift=11pt] {Checkpoint};
\node[io, anchor=south] (sin2) at ([xshift=0in]spanel.north) [yshift=11pt] {Tune file};
\node[io, anchor=south] (sin3) at ([xshift=0.85in]spanel.north) [yshift=11pt] {RAM budget (weights)};
\node[io, below=11pt of spanel] (sout) {Generated tokens};
\draw[->] (sin1) -- (sin1|-spanel.north);
\draw[->] (sin2) -- (sin2|-spanel.north);
\draw[->] (sin3) -- (sin3|-spanel.north);
\draw[->] (s1) -- (s2);
\draw[->] (s2) -- (s3);
\draw[->] (spanel) -- (sout);
\end{tikzpicture}
\caption{Overview of MINE. Left: tuning, done once for a given checkpoint and GPU; right: inference with the MINE plugin in vLLM.}
\label{fig:overview}
\end{figure}

We give an overview of MINE and how it is integrated into the existing vLLM engine \citep{vLLM23} to efficiently serve entropy-coded checkpoints (model in a specific format/quantization) under a memory budget. MINE consists of two parts, as shown in Figure~\ref{fig:overview}: the tuning (left), which is done once for a given checkpoint and GPU, and the actual inference (right).

\paragraph{Tuning.}
For the tuning, we first enumerate all (feasible) codings over some design choices: combining 1/2/4 weights into a symbol to code, the number of columns $s_c$ to process per thread, the size of our decoding tables $M$, and how to represent \texttt{FP8} values in the decoding tables. Codings that do not yield a compression in expectation are pruned. Then, for some data types, we search the number of values $t$ to entropy-code (the others are escaped and stored in full precision). We encode using dtANS-LLM and tune how many warps to put in a block to minimize cold-cache \texttt{mvm} latency. We compute the runtime vs.~model size Pareto front and, for each point on it, also tune warps per block for the other kernels \texttt{dequant} / \texttt{encode} / \texttt{decode}. Since the described search space is still vast, we will show later how we further prune it to keep tune times manageable and runtimes competitive. The final output is collected in a tune file.

\paragraph{Inference.}
For the inference, MINE gets as input a checkpoint together with the computed tune file and the RAM budget. The RAM budget applies to the weights only, so the user has to leave some headroom for KV cache and materialized tensors/activations for prefill. MINE then searches the runtime vs.~model size Pareto front for the fastest combinations of codings for all kernel launches under the budget. This is a multiple-choice knapsack problem, which we solve using dynamic programming \citep{Knapsack04}. Then, it loads the uncompressed tensors to the GPU one by one and encodes them. Single-stream decoding (batch size 1) is then served by \texttt{mvm}, and for all other batch sizes, \texttt{dequant} materializes the weights to global memory followed by a GEMM which might benefit from L2 cache.\footnote{For \texttt{E4M3} we \texttt{decode} instead and call the corresponding vLLM kernel.}

\section{Background: Fast Fused Entropy Decoding}

We start this section by briefly explaining the main aspects for efficient GPU inference. Then, we give a high-level explanation of the entropy coding methods dtANS-LLM builds on. Entropy coding describes the idea of compressing a stream of symbols from an alphabet $\Omega$ using close to $H(P)$~bits per symbol where $H(P)$ is the Shannon entropy of the distribution $P$ of the symbols in the stream. $H(P)$ is a lower bound for the size of the compressed stream \citep{Shannon48}.

\paragraph{Efficient GPU Code.} On NVIDIA GPUs, instructions are issued to \emph{warps} of 32 \emph{lanes}. For memory accesses to global memory to be efficient, neighboring threads have to access consecutive addresses (coalescing). CUDA also exposes fast on-chip \emph{shared memory} which is organized in 32 banks. In each cycle, each bank can provide one 32-bit \emph{word}, and requests causing a \emph{$c$-way bank conflict} are serialized into $c$ transactions (with 2-way conflicts being free on consumer GPUs for 32-bit accesses \citep{FreeBankConflicts23}). Lastly, warps are grouped into \emph{blocks} which are then executed on \emph{streaming multiprocessors} (SMs). Each SM can execute as many blocks (and thus threads) as the tightest limit over several limited resources allows: threads, registers, and shared memory. The \emph{occupancy} describes the resulting number of threads/warps resident on any SM in parallel \citep{ProgrammingGuide25}.

\paragraph{tANS.}
\emph{Tabled Asymmetric Numeral Systems} (tANS) \citep{ANS13} is an entropy coding method that achieves rates close to $H(P)$ with reasonable computational cost. The name comes from the existence of a \emph{decoding table} with $M = 2^m$ entries, each containing one of the occurring symbols, a new state $x'$, and a number $k$ of bits to load from a compressed bit stream $v$. The decoder maintains a state $x$. It uses the last $m$~bits of $x$ to perform a look-up into the decoding table which yields the decoded symbol and the information needed to construct the next state $x = x' \ll k~|~v_k$ where $v_k$ are the next $k$~bits of $v$. Encoding works by simulating the actions of the decoder in reverse. The encoded representation consists of a bit stream and a decoding table.
The two main issues toward a fast GPU implementation are the dependencies between decoding subsequent symbols (limited instruction-level parallelism (ILP)) and the need to load a variable number of bits from each thread after each symbol (uncoalesced accesses) \citep{dtANS26}.

\paragraph{dtANS.}
\citet{dtANS26} alleviated these issues by splitting the decode and state reconstruction. They use tables of size $M=2^{12}$ and obtain the table indices by rearranging a state of three 32-bit words into an \emph{octet} of 12-bit indices. Then, they can decode the symbols in parallel and obtain 8 \emph{digit/base pairs}, which can be appended to some state $x$ using Mixed Radix Numeral Systems. This increases ILP. Then, as many full 32-bit words as possible are extracted from $x$. As dtANS limits the base in the table to $2^8$, $x$ can grow by at most $2^{64}$ and at most 2 words can be extracted. Words that cannot be extracted are \emph{conditionally} read from a stream in global memory instead. For these reads to be coalesced, intra-warp communication is required. The third word needed for the next round is always loaded \emph{unconditionally}. Limiting the base to $2^8$ causes each symbol to compress to at least $12 - 8 = 4$~bits, which limits the achievable rate for very skewed distributions (i.e., when some symbol has probability higher than $2^8/2^{12}=6.25\%$).

\section{MINE for Fast Single-Stream Inference}
\label{sec:mine}

\begin{figure}[t]
    \centering
    \includegraphics[width=\linewidth]{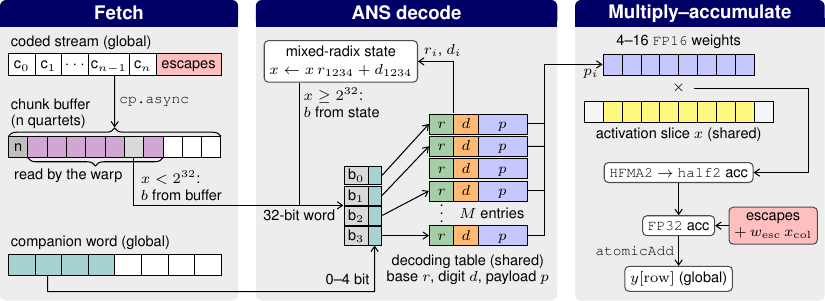}
    \caption{Overview of the performance-critical \texttt{mvm} kernel. Compressed data is fetched and decoded into weights which are used in place for fused matrix-vector multiplication with activations.}
    \label{fig:dtANS_LLM}
\end{figure}

In this section, we present MINE for fast single-stream inference. We explain our optimized entropy coding method dtANS-LLM and how we obtain the parameters for optimal single-stream decoding performance under memory constraints. Figure~\ref{fig:dtANS_LLM} visualizes the performance-critical \texttt{mvm} kernel.

\paragraph{Grid Layout.}
We split each tensor into $s_r\times s_c$ tiles. Each tile is mapped to one warp where each lane processes one row. Tiles are grouped into blocks vertically, such that blocks share input dimensions and can cache their activations. The data type used for activations follows the vLLM baseline for \texttt{FP8} and \texttt{BF16} but is \texttt{FP16} for the quantized checkpoints (vLLM uses \texttt{BF16}).

\subsection{dtANS-LLM}

We describe how we fuse an adapted dtANS \citep{dtANS26} into the kernels efficiently.

\paragraph{Fetching Data.} dtANS uses intra-warp communication to load the stream from global memory in coalesced form. As this approach suffers from high latency, we introduce a shared memory buffer which we explain following the left block of Figure~\ref{fig:dtANS_LLM}. The buffer comprises a \emph{chunk} of 1~KiB per warp and is copied using \texttt{cp.async}. The first word of every chunk counts after how many iterations $n$ the next chunk has to be loaded (takes the loop iteration off the critical path), upon consumption the remaining data is partially flushed into the lanes (remaining lanes read from the next chunk), and the metadata ensures that only the required part of the last chunk is copied.

\paragraph{Decoding Tables.}
dtANS-LLM uses the \emph{merged} 32-bit table design from dtANS which reserves 8~bits for a digit and base each such that 16~bits remain for the \emph{payload}. In contrast to dtANS, it stores the base directly instead of its decremented counterpart which limits symbols to 255 slots and thus loses up to $\log_2(256/255)\approx 0.006$~bits for each symbol that would have used 256 in an optimal approximation. While dtANS demonstrated how to use the decremented value without additional instructions, it did not consider the critical path which is shorter for trivial accumulation.

\paragraph{Data Types.}
dtANS-LLM is designed to support a variety of payloads. For now, these are \texttt{BF16}, \texttt{FP8}, \texttt{FP4}, and \texttt{int2} / \texttt{int3} / \texttt{int4}. Out of these, only \texttt{BF16} requires specialized handling because a decoding table for all $2^{16}$ values would be too large for shared memory. Given that the 1-bit sign and 7-bit mantissa bits are distributed almost uniformly and independently of the 8-bit exponent, we decided to follow the established idea of only encoding the exponent and storing sign and mantissa as a plain byte \citep{ZipNN25}. In our own experiments, we also noticed that the distributions of \texttt{FP4} and \texttt{int2} / \texttt{int3} / \texttt{int4} are quite uniform, while those of \texttt{FP8} and the exponent of \texttt{BF16} are very skewed (many values appear very infrequently if at all).

Given that our table entries support 16-bit payloads, we could combine up to 8 \texttt{int2}, 5 \texttt{int3}, 4 \texttt{int4} / \texttt{FP4}, and 2 \texttt{BF16} / \texttt{FP8} values in a single symbol. Observe how constructing $k$-tuples over $t$ values calls for $M \geq t^k$ to make every combination encodable. In practice, to also get a good rate, we need a multiple of that. As tables are stored in shared memory, they should not be too large, and we limit \texttt{int2} to Quad mode ($k=4$) and \texttt{int3} / \texttt{int4} / \texttt{FP4} to Pair mode ($k=2$). These limits also enable payload designs that allow for efficient unpacking.

\paragraph{Escapes.}
By the same logic, \texttt{BF16} and \texttt{FP8} are effectively limited to Single mode ($k=1$). However, as many symbols are rare, we restrict the decoding table to the $t$ most frequent symbols. This enables encoding 2 values per symbol if $t^2 \leq M$ but requires a mechanism to handle values that are not in the table. dtANS resolves this issue with a special escape symbol that causes the decoder to read a full-precision value, which in turn requires communication and memory transactions. Differently, dtANS-LLM encodes a 0 instead and adds the actual value to a sparse correction stream. This reconciles fast processing of the dense matrix and the rare values. The correction stream is appended directly to the ANS stream and contains one 32-bit \emph{record} for each escape: a 16-bit weight, an 11-bit column index (relative to the start of the warp's slice), a 1-bit flag (whether the current escape is the last for the row), and a 4-bit counter (for how many rounds to pause before processing the record). Each lane processes the escapes of its assigned row, and memory accesses are coordinated as the conditional loads were before. Thus, the procedure takes as many steps as there are escapes in the row with the most records. In practice, escapes are sufficiently random and thus spread. The 4-bit counter is straightforwardly used to space out reads equally, which gives good performance in practice. We tried more advanced scheduling but it yielded no measurable returns. 

\paragraph{Efficient Innermost Loop.} 
In dtANS, storing packed indices and words for the next iteration requires 3 registers each. For different $M$, this is aggravated: $M=2^9$ requires 9 each (lowest common multiple of 9 and 32 is $9 \times 32$) and $M=2^{10}$ requires 5. As dtANS-LLM should support multiple $M$ to cater to the different data types, we simplify the innermost loop as shown in the middle block of Figure~\ref{fig:dtANS_LLM}. Our design only takes 8~bits from the ANS state and the remainder from an uncompressed companion stream. The ANS state now only covers a single word and each iteration only decodes a \emph{quartet}. If the number of bits read from the companion stream is a power of two, unrolling the innermost loop such that an entire word is consumed reduces overhead drastically. Hence, we support $M \in \{2^8, 2^9, 2^{10}, 2^{12}\}$ with 0, 1, 2, and 4~companion bits, requiring unrolling of 0, 8, 4, and 2 quartets, respectively. Our scheme has the additional advantage of having less than one word of information in the internal state at the end of the loop whereas dtANS had at most two (this information cannot be used anymore and is thus a constant overhead).

\paragraph{Fused \texttt{\textbf{mvm}}.} The \texttt{mvm} kernel required for fast single-stream inference fuses a dense matrix-vector product of a layer with an activation vector into the decoding. A visualization can be found in the right block of Figure~\ref{fig:dtANS_LLM}. It is highly optimized, accumulates 64 weight-activation products (but at most all of a group) in \texttt{FP16} before flushing to \texttt{FP32}, unrolls the innermost loop such that the flush and handling of the companion stream do not have to track state, and for quantized checkpoints with row-wise scales/zeros dequantizes efficiently per group/slice in an \emph{epilogue} rather than per weight.

\subsection{Auto-Tuning}
In this section, we explain how MINE chooses the hyper-parameters for dtANS-LLM that obtain optimal performance for single-stream decoding, and thus the \texttt{mvm} kernel is the only runtime tuning objective. 
We do not only store the fastest coding but compute the speed--size Pareto front to later find the optimal coding under memory constraints (smaller codings might be more computationally expensive to decode). We first explain tuning for the generic case and then illustrate the special handling for the sparse alphabets of \texttt{FP8} and \texttt{BF16}. MINE also contains a \texttt{dequant} kernel that materializes the layer to global memory, and \texttt{encode} / \texttt{decode} kernels which are not on the critical single-stream decode path. These additional kernels are only tuned on the points that made it to the Pareto front (with the details being omitted for brevity).

\paragraph{Compressed Size Prediction.}
In order to prune the search space and tune efficiently, we will require accurate estimates of the compressed size. To this end, we use the fact that the compressed size can be closely predicted by the cross-entropy between the distribution and its approximation by the tables. It can be computed efficiently from a histogram (which we keep in a cache) and is a lower bound on the actual compressed size because dtANS-LLM adds an almost constant overhead. Thus, its argmin is close to the argmin of the compressed size.

\paragraph{Launch.}
In most settings, the computational graph of an LLM exposes parallelism to some degree, allowing multiple matrix-vector products to be computed in parallel. Thus, grouping these together to share a single kernel \emph{launch} can significantly reduce overhead, increase device utilization, and improve latency. For simplicity we require these \emph{sections} in a launch to have the same input dimension and quantization method/parameters. For the default transformer architecture, there are 4 launches per layer: q/k/v, gate/up, o\_proj, and down\_proj. To keep our search space over \texttt{mvm} kernels small, guarantee similar work per block, and use resources optimally (mixing pays the maximum of each resource), we choose mode, $s_c$, and $M$ per launch. Each combination constitutes a \emph{coding}, and which we evaluate depends on the launch. First, modes are chosen depending on the data type: Single and Pair are supported by all and Quad only by \texttt{int2}. Second, $s_c$ are chosen from $\{64,128,256,512,1024,2048\}$ such that they divide the input dimension, allow sufficient unrolling, and are a multiple of the group size (if applicable). By choice, $s_c$ is also a multiple of the 64 weight-activation products that are accumulated in \texttt{FP16} before being flushed to \texttt{FP32}. Third, $M$ are taken from $\{256,512,1024,4096\}$ in order until our lower bound increases by more than 1\% over the minimum observed until then (larger tables require more shared memory and computation, so they must pay for themselves in smaller size). If vLLM kernels are available, we add their runtime and size to the Pareto front and skip codings whose lower bound exceeds the uncompressed size.

\paragraph{Tuning Grid.}
For each coding, we want to find the optimal number of blocks per SM and warps per block. We start by computing for each number of blocks per SM how many warps each block can have to still fit in shared memory and the threads per SM limit. Then, we take the maximum number of warps per block that still compiles without register spilling when we provide nvcc with the corresponding launch bounds (we can maintain a cache of compiled binaries). When serving, the cache will be cold and thus we need accurate cold-cache measurements for the time taken by the launches. To this end, we reproduce the kernel as many times as necessary to fill the L2 cache at least twice. Then, we capture the compute graph and take the median runtime of 5 bursts, each of which comprises as many replays as required to spend 400~$\mu$s. The data is then added to the front.

\paragraph{Tuning $\boldsymbol{t}$.}
For \texttt{FP8} and \texttt{BF16} there is another axis to tune: the number $t \leq 256$ of weights/exponents to store in the decoding tables. Values that are not stored in the table are escaped and require an additional 4-byte record. By entropy, it is optimal to keep the $t$ most frequent codes (and always 0 as we need it for escapes). We can adapt our lower bound on the compressed size accordingly by adding the size of the records (we already have the necessary histogram). Generally, we must assign at least one table slot for each payload tuple, so $t \leq M$ for Single and $t^2 \leq M$ for Pair mode. As $M \geq 256$, this is trivially fulfilled for Single and for Pair we require $t \leq 16, 22, 32, 64$ for $M = 256, 512, 1024, 4096$, respectively. Since good choices of $t$ depend on the distribution of weights in the tensor, we search it per section but in parallel per launch (to keep the search space linear). So for each coding, we first use the lower bound to find an estimate of the byte-optimal $t$ and tune the warps per block for them. The expected runtime of \texttt{mvm} depends on two aspects: the amount of memory to transfer (increases for smaller and larger $t$) and compute (decreases with $t$). Thus, reducing $t$ would increase compute without reducing transfer size and consequently is not expected to win. In contrast, increasing $t$ will decrease compute while possibly increasing transfer size, pushing the kernel to the memory-bound regime (left in a roofline plot \citep{Roofline09}). Thus, for each section, we compute the escapes-saved-per-additional-byte ratio when incrementing $t$ (while $t$ can still be increased). Then, we increment the $t$ with the highest ratio until the memory transfer size is about to increase by more than 1\% (or all $t$ are at the cap), tune that selection, and iterate. As before, if the lower bound is above the plain tensor size, we stop considering the coding (which might already happen at minimal $t$). Lastly, we bisect runtime gaps larger than 1\% (on top of measurement noise) to get a smooth front.

\paragraph{Floats.}
Lastly, for \texttt{FP8}, we tune two payload layouts for Pair (Single stores \texttt{FP16}): \texttt{E4M3} which has to be converted to \texttt{FP16} and \texttt{E5M2} which can simply be used as the high byte of \texttt{FP16} \citep{FP8_22}. The former has a constant overhead while the latter requires escaping values that are not representable. In practice, the latter works well only for EntQuant \citep{EntQuant26} because we observe that most codes lie at the subnormal floor, which can be represented in \texttt{E5M2}.

\section{Related Work}
Various works have explored schemes to quantize the weights of LLMs as discussed in Section~\ref{sec:introduction}. To date, far fewer have considered lossless entropy coding due to inherent overhead. Decompressing into a global buffer \citep{NeuZip24,DFloat11_25,EntQuant26} yields good compression but negative speedups for single-stream inference due to the additional latency. Some works use the freed RAM for larger batch sizes to increase throughput \citep{ShannonBound26,NF12_26}. Other approaches employ domain-specific techniques to obtain good compression and little speedup \citep{ZipServ26,NF12_26}. In Section~\ref{sec:evaluation}, we compare MINE in the challenging single-stream decoding setting to the four of these works with source code available.

There are also works employing entropy coding to (de)compress to/from global memory \citep{ParallelHuffman18,ParallelANS19,GapArrays2020,DietGPU22,Recoil23,nvCOMP} and shared memory \citep{nvCOMPDx26}. Recently, a fast fused entropy decoding method has been proposed for another workload \citep{dtANS26}. There have also been considerations to implement entropy coding in hardware \citep{E2MC17,Huff-LLM25,Ecco25}.

\section{Evaluation}
\label{sec:evaluation}
We evaluate MINE on Llama~3.1~8B~Instruct \citep{Llama3_24} in 19 formats/quantizations and on two GPU cards. The model has a convenient size and fits one card uncompressed but not the other. First, we show an extensive steady-state single-stream decoding throughput comparison between MINE, vLLM \citep{vLLM23,Marlin25,Humming26}, and related works \citep{DFloat11_25,NeuZip24,EntQuant26,ZipServ26}. Then, we contrast these results with the prefill performance (for which MINE is not optimized).

\paragraph{Experimental Setup.}
We conducted our experiments on two GPUs: an RTX~4060~Ti / RTX~5090 with 16 / 32~GB of VRAM, 32 / 96~MB L2 cache, and 34 / 170 streaming multiprocessors (SMs) with 128~KB L1 cache each (of which up to 100~KB can be used as shared memory). We highlight that the former card has 2.64 ALUops/bandwidth-bit compared to 2.21 (19\% more).\footnote{We consider three-operand operations that have a throughput of 64 per SM per cycle. Further, we used frequencies measured during our actual workload: 2790~MHz on the 4060~Ti and 2917~MHz on the 5090.} They were put in the same machine with an Intel Xeon w5-2465X with 33.75~MB L3 cache and 128~GB DDR5 ECC RAM. As an operating system, we installed Ubuntu 24.04.4 LTS with kernel 7.0.0-31-generic and NVIDIA driver 595.84 (open). MINE's kernels were compiled via torch using nvcc 13.1.115, with flags \texttt{-O3 --use\_fast\_math} and pinned architecture \texttt{-gencode} \texttt{arch=compute\_89,code=sm\_89} / \texttt{arch=compute\_120,code=sm\_120}.

\paragraph{Checkpoints.}
We evaluated MINE on 19 checkpoints of Llama~3.1~8B~Instruct: \texttt{BF16}; \texttt{E4M3} \citep{FP8_22}; \texttt{MXFP4} \citep{MXFP4Spec23,MXFP4Eval23}; \texttt{NVFP4} \citep{NVFP4_25}; EntQuant with 2.1 / 3.0 / 3.9~bits of entropy per weight \citep{EntQuant26}; asymmetric RTN with 2~bits (groups of 64) and 3 / 4~bits (groups of 128) \citep{RTN23}; GPTQ with groups of 128 and 2 (asymmetric), 3 (asymmetric), and 4 (symmetric) bits \citep{GPTQ23}; symmetric AutoRound with 2~bits (groups of 128 / 64 \citep{AutoRound25}) and 4~bits (groups of 128) \citep{AutoRound24}; HQQ with 2~bits (groups of 32 in column dimension) and 4~bits (groups of 64) \citep{HQQ23}; and asymmetric AWQ with 4~bits (groups of 128) \citep{AWQ24}. The \texttt{lm\_head} is always stored in \texttt{BF16} but compressed as well. We also considered EXL3 \citep{EXL3_25} based on \citet{QTIP24} and GGUF \citep{GGUF23} i-quants \citep{IQuants24}, but they would have required specialized fused kernels and did not exhibit any entropy headroom (so tuning would most likely have picked the default kernels). Tune times ranged from 14.3~minutes (HQQ, 4 nominal bits, groups of 64; RTX~4060~Ti) to 262.3~minutes (EntQuant, 2.1~bpw; RTX~5090). Generally, \texttt{FP8} checkpoints take the longest due to the two payload formats and the additional search dimension over $t$ (especially if there are many escapes).

\paragraph{Steady-State Decode Throughput.}

We evaluate MINE's steady-state single-stream decoding throughput on the 19 checkpoints and compare it to performance-optimized baselines where available: vLLM~0.27.1 \citep{vLLM23} with its cuBLAS \citep{cuBLAS} / CUTLASS \citep{CUTLASS}, Marlin \citep{Marlin25}, and Humming \citep{Humming26} kernels. For HQQ, we did not find any high-performance baseline. We also include memory-optimized baselines that have source code available: DFloat11 \citep{DFloat11_25}, ZipServ \citep{ZipServ26}, and two engines which use nvCOMP \citep{nvCOMP} for entropy coding: NeuZip \citep{NeuZip24} and EntQuant \citep{EntQuant26}. All of these required a different software stack configuration which we omit for brevity. The throughput was determined using the runtime difference between a 64- and 128-token generation following a random 512-token prompt. The runtimes were measured as the median of 3 runs. 

\begin{figure}[ht]
    \centering
    \includegraphics[width=\linewidth]{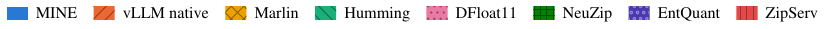}\\[4pt]
    \begin{subfigure}{0.465\linewidth}
        \centering
        \includegraphics[width=\linewidth]{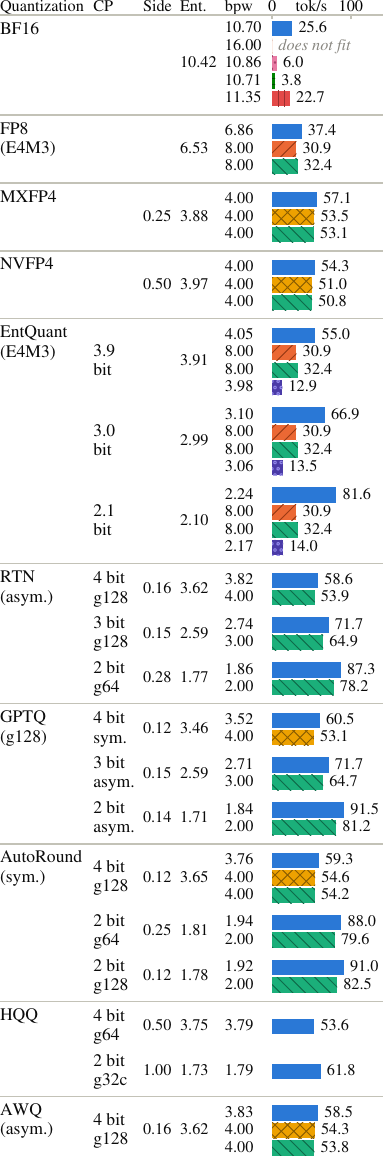}
        \caption{RTX~4060~Ti (16~GB)}
        \label{fig:throughput_4060_ti}
    \end{subfigure}
    \hfill
    \begin{subfigure}{0.465\linewidth}
        \centering
        \includegraphics[width=\linewidth]{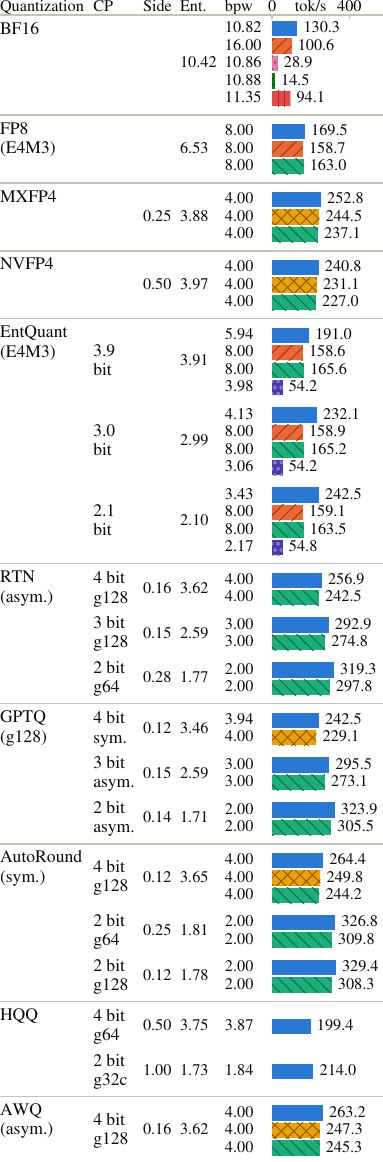}
        \caption{RTX~5090 (32~GB)}
        \label{fig:throughput_5090}
    \end{subfigure}
    \caption{Evaluation of MINE (blue) and other inference engines on checkpoints of Llama~3.1~8B~Instruct and two GPUs (left and right half). Columns state quantization method, checkpoint (CP), size of dequantization side streams, entropy bits per weight, achieved bits per weight, and single-stream (batch 1) steady-state decode throughput.}
    \label{fig:throughput}
\end{figure}

In Figure~\ref{fig:throughput} we summarize the results. For each quantization method, we list the checkpoints (CPs), and for each checkpoint the inference engines that serve it. Then, starting from the left, we give the size of dequantization side streams, entropy bits per weight, achieved bits per weight, and single-stream (batch 1) steady-state decode throughput. The color of the bar corresponds to the engine.

We highlight two key observations. First, MINE achieves the best throughput on all checkpoints. This is expected behavior because MINE can pass layers on which it cannot improve to vLLM. The maximum speedup over the fastest competitor is 2.52$\times$ / 1.48$\times$ and is achieved on EntQuant with 2.1~bits. As expected, this is the checkpoint with the most relative entropy headroom and we highlight that MINE's throughput is on par with conventional 2-bit quantization schemes on the RTX~4060~Ti. Generally, the true entropy coding baselines (DFloat11, NeuZip, and EntQuant) are at least 3.52$\times$ slower than MINE. Only ZipServ shows comparable performance. While making strong assumptions about the distribution of values, achieving slightly worse bpw, and being limited to \texttt{BF16}, it only achieves 0.88$\times$ / 0.72$\times$ MINE's throughput. However, it builds on vLLM~0.9.1, which reaches 91.8~tok/s on the RTX~5090, resulting in a 2.6\% speedup.

Second, MINE prefers smaller codings on the RTX~4060~Ti compared to the RTX~5090. This is because the RTX~5090 is less memory-bound, favoring parameters that lead to less work over those that lead to less memory transfer. Further, more launches fall back to vLLM because the RTX~5090 has more SMs that reduce the number of waves and make latency more critical (dtANS-LLM has to transfer an entire chunk first). For many checkpoints, MINE never chooses its own kernels on the RTX~5090 and its speedups result exclusively from combining native/Marlin/Humming kernels and compressing the \texttt{lm\_head}. Note that MINE could still compress in these cases to free up memory but would suffer from a throughput penalty.

\paragraph{Prefill.}
As \texttt{mvm}, which is crucial for single-stream decoding, is the main optimization goal and only implemented for batch size 1, larger batch sizes (for prefill) have to be handled by \texttt{dequant}/\texttt{decode} and a GEMM. For instance, the latency of materialization is only 1.3--2.9$\times$ that of \texttt{mvm} and the subsequent GEMM at batch size 2048 is another 14.7--20.2$\times$ (on the launches of Llama~3.1~8B~Instruct in \texttt{BF16}). This overhead of 8.1--15.4\% is partially compensated by the materialized weights being L2 resident and thus prefill reaches 12,860~tok/s compared to vLLM's 13,650~tok/s (94\%, 4096-token prompt on the RTX~5090; the RTX~4060~Ti has no uncompressed reference).

\section{Conclusion}

We showed with MINE that fast entropy decoding can be used to accelerate inference even in the challenging single-stream decoding setting. At the same time, MINE can adapt to tighter memory budgets, trading performance for better compression and thus fit larger models into less RAM. Thus, it would be promising to extend MINE beyond single-stream decoding where optimization objectives become more heterogeneous: fused kernels for small batch sizes to avoid the full cost of materialization, compressing the KV cache, and switching codings dynamically as the optimal coding might depend on the workload (e.g., between prefill and decode).

Lastly, our work motivates a new metric for quantization methods. As MINE can store weights close to their entropy and provide inference at equivalent speeds as well, the entropy of the quantized weights gets more important than their nominal rate and finding quantization schemes with low entropy becomes more important. 

\clearpage
\section{AI Use}
We used generative AI to \emph{implement methods} (coding) and \emph{design or provide feedback on research methodology or experiments} (evaluation). We also used it to \emph{create or modify scientific figures or images} (all figures and plots), \emph{suggest experimental parameters} (identify hyper-parameters to tune), \emph{create or edit software code}, \emph{creation of artifacts} (reproducibility artifact, statement, and appendix), \emph{draft parts of a research paper} (abstract and reproducibility statement and appendix), \emph{identify relevant literature}, \emph{summarize or analyse existing literature}, \emph{sourcing/searching for information}, \emph{edit a research paper to improve readability} (also finding disagreements between paper and code), \emph{format references} (we checked all details manually), and \emph{propose a title or keywords for a research paper}.

\section{Reproducibility Statement}
The reproduction package is the single archive \texttt{mine-artifact.zip}. It contains the MINE source code (codec, CUDA kernels, auto-tuner, and vLLM plugin) and one numbered script per experiment, and it downloads everything else: third-party code at pinned commits, Python packages from exact lock files, and the model and datasets at fixed Hugging Face revisions. Appendix~\ref{app:repro} maps Figure~\ref{fig:throughput} and every number of Section~\ref{sec:evaluation} to the script that produces it, describes how to serve MINE under a memory budget, and lists the software versions, the recipes of all quantized checkpoints, the measurement protocols, the hardware requirements, and all changes we made to third-party code. The dtANS-LLM format and the tuning procedure are specified in Section~\ref{sec:mine}, the hardware and software environment in the Experimental Setup paragraph of Section~\ref{sec:evaluation}. All scripts detect the GPU at runtime, so the complete pipeline runs on any NVIDIA GPU with compute capability 8.9 or higher and at least 16~GB of memory (24~GB to also run the baselines that need the uncompressed \texttt{BF16} model on the GPU); since MINE tunes for the GPU it runs on, absolute throughput is only expected to match ours on the same GPU models.

\clearpage
\bibliography{iclr2027_conference}
\bibliographystyle{iclr2027_conference}

\clearpage
\appendix
\section{Reproducing the Experiments}
\label{app:repro}

The reproduction package is a single archive, \texttt{mine-artifact.zip}, that reproduces all experiments of Section~\ref{sec:evaluation} from scratch. It holds only our code: the MINE package (codec, CUDA kernels, auto-tuner, vLLM plugin) and the scripts that drive the experiments. The setup scripts download all other code at pinned commits and apply our changes (\texttt{setup/fetch.sh}), build the Python environments from exact lock files (\texttt{setup/envs.sh}), and download the model, the DFloat11 checkpoint, and the evaluation data at fixed revisions (\texttt{setup/download.sh}). The numbered steps of Table~\ref{tab:repro-map} are then run in order; every step is resumable, and the package's \texttt{README.md} gives the commands, the requirements, and a quick check on a single checkpoint.

\begin{table}[h]
\centering\small
\caption{Results of the paper and the steps of \texttt{mine-artifact.zip} that produce them. Steps \texttt{10}--\texttt{15} run once per GPU and store their results under a label derived from the GPU's name; step \texttt{20} needs no GPU.}
\label{tab:repro-map}
\begin{tabular}{p{0.56\linewidth}p{0.36\linewidth}}
\hline
Result & Step \\
\hline
The 18 quantized checkpoints & \texttt{run/01\_checkpoints.sh} \\
Entropy $H(P)$; \texttt{BF16} perplexity (reference of ZipServ's gate) & \texttt{run/02\_stats.sh} \\
Tune files and tune times & \texttt{run/10\_tune.sh} \\
Decode throughput of MINE, vLLM (native / Marlin), and Humming & \texttt{run/11\_serve.sh} \\
DFloat11, NeuZip, EntQuant (nvCOMP), ZipServ, and its vLLM~0.9.1 & \texttt{run/12\_baselines.sh} \\
Prefill paragraph & \texttt{run/13\_prefill.sh} \\
GPU clocks under load (Experimental Setup) & \texttt{run/15\_clocks.sh} \\
Figure~\ref{fig:throughput} (side streams, $H(P)$, bpw, tok/s) and every number quoted in Section~\ref{sec:evaluation} & \texttt{run/20\_figures.sh} \\
Serving under a memory budget (Figure~\ref{fig:overview}; optional, no reported result) & \texttt{run/14\_budget.sh} \\
\hline
\end{tabular}
\end{table}

\paragraph{Serving under a memory budget.} The evaluation serves every checkpoint with its fastest tuned selection. \texttt{run/14\_budget.sh} serves MINE instead with the fastest selection whose weights fit a given budget, as in the right half of Figure~\ref{fig:overview}: from the per-launch codings in the tune file, \texttt{mine/tools/pick\_budget.py} chooses one per launch so that the summed batch-1 time is minimal while the weights (coded linears, embeddings, norms, and all other resident weights) fit the budget, a multiple-choice knapsack that it solves by dynamic programming and certifies against a lower bound. The selection is passed to the plugin in \texttt{MINE\_TUNE\_SELECT} and served with the decode protocol below. The budget is given in bytes or as a fraction of the fastest selection's footprint, and may lie anywhere between the smallest and the fastest selection; the package's \texttt{README.md} gives the commands.

\paragraph{Software.} MINE runs as a plugin of vLLM~0.27.1 (commit \texttt{6e448d0}, built from source) with PyTorch~2.13.0 (CUDA~13.0 build); its kernels are JIT-compiled for the running GPU by nvcc~13.1 with \texttt{-O3 --use\_fast\_math}. The vLLM baselines run in a separate environment with the official vLLM~0.27.1 wheel and without the plugin; Humming is the \texttt{humming-kernels}~0.1.10 package vLLM~0.27.1 requires, enabled with \texttt{--quantization humming}. The checkpoints were produced with GPTQModel~7.3.4 (\texttt{fe8aecb}), auto-round (\texttt{f17d9cd}), llm-compressor (\texttt{fc2cf94}) with compressed-tensors~0.18.1a20260806, HQQ (\texttt{d88a488}), and EntQuant (\texttt{a957fec}). The baselines are DFloat11~0.5.0 with its published Llama~3.1~8B checkpoint (revision \texttt{8fcadb8}), NeuZip (\texttt{1224c60}) with nvCOMP~4.1.1.1, EntQuant's decoder with nvCOMP~5.3.0.16, and ZipServ (\texttt{f69cea9}) on its vLLM~0.9.1 fork (vLLM \texttt{7b1895e}) with the GEMM kernel of the ZipServ artifact (\texttt{6f8a209}). The model is Llama-3.1-8B-Instruct at revision \texttt{0e9e39f}; WikiText-2, C4, and pile-10k are pinned to revisions \texttt{b08601e}, \texttt{1588ec4}, and \texttt{127bfed}.

\paragraph{Checkpoints.} All checkpoints quantize the 224 decoder linears and keep embeddings and \texttt{lm\_head} in \texttt{BF16}. \texttt{E4M3}: llm-compressor's \texttt{FP8\_DYNAMIC} scheme (weights per channel, activations dynamically per token); \texttt{MXFP4}/\texttt{NVFP4}: its weight-only \texttt{MXFP4A16}/\texttt{NVFP4A16} schemes. EntQuant: its entropy-regularized \texttt{E4M3} quantizer, data-free, with $(\lambda, \text{lr}) = (58, 0.25)$, $(14.5, 1.0)$, and $(3.9, 1.0)$ for 2.1, 3.0, and 3.9~bits. RTN: auto-round with zero iterations (which, in this version, applies its importance-weighted RTN on the calibration set), asymmetric, GPTQ format. GPTQ: GPTQModel with damping 0.05 and sequential quantization; activation reordering (\texttt{desc\_act}) for the symmetric 4-bit checkpoint, GPTQModel's default group-aware reordering for the others. AutoRound: 200 iterations with \texttt{--enable\_alg\_ext}, symmetric. AWQ: llm-compressor's \texttt{W4A16\_ASYM} AWQ recipe. HQQ: 4~bits with groups of 64 along rows and 2~bits with groups of 32 along columns, 16-bit scales and zeros, data-free. GPTQ and AWQ calibrate on 128 random windows of 2048~tokens from one C4 training shard; RTN and AutoRound on auto-round's default of 128 samples of 2048~tokens from pile-10k. The calibrated methods are not bit-reproducible across GPUs and driver versions; the others are deterministic. For serving, the scripts convert without changing any value: HQQ's pickle to safetensors (MINE), EntQuant to a compressed-tensors \texttt{FP8} checkpoint (vLLM, Humming), and the zero points of the asymmetric RTN and GPTQ checkpoints from the GPTQ~v1 to the v2 convention (Humming). Humming is shown for the checkpoints on which it produced correct output in our runs; the scripts serve exactly these.

\paragraph{Measurement protocols.} \emph{Decode} (MINE, vLLM, Humming, ZipServ, and its vLLM~0.9.1): a random 512-token prompt followed by exactly $G$ greedy tokens (EOS ignored) for $G = 64$ and $G = 128$, CUDA graphs enabled, prefix caching disabled, median of 3 runs after one warm-up; the steady-state rate $1/t_\text{dec}$ is solved from $t(G) = t_\text{pre} + G\,t_\text{dec}$. DFloat11, NeuZip, and EntQuant (nvCOMP) run under their own HF~Transformers stacks with the same prompt and estimator, with three interleaved repetitions of both $G$ after two warm-ups. Bits per weight are measured over the decoder linears for every engine (for DFloat11 from its published checkpoint; for NeuZip on each GPU, as nvCOMP's compressed size differs between GPUs). \emph{ZipServ's gate}: ZipServ is timed only after its WikiText-2 perplexity (non-overlapping windows of 2048~tokens, all positions scored, prompt log-probabilities of its vLLM) was within $\pm 0.5\%$ of the \texttt{BF16} model under HF~Transformers, and its greedy batch-1 tokens agreed with the same vLLM's dense \texttt{BF16} path (at least 90\% of the tokens of 8 prompts and none of the first 8, or else every divergence a near-tie of dense \texttt{BF16}: ZipServ's token in its top 2 with a margin of at most 0.25~nats). \emph{Tuning}: every launch is timed cold-cache (copies covering $2\times$ the L2 cache, CUDA-graph replays, median of 5 bursts of at least 400~$\mu$s); the \texttt{BF16} \texttt{lm\_head} is identical in all checkpoints and is tuned once per GPU, outside the reported tune times. \emph{Clocks}: the median SM clock over all 100~ms samples with at least 90\% utilization during a MINE decode run.

\paragraph{Hardware requirements and portability.} Nothing in the package is specific to our GPUs: the GPU is chosen by its index, and the label, architecture, JIT target, occupancy limits, and L2 size are read at runtime. MINE requires compute capability 8.9 or higher (its \texttt{FP8} conversion instructions and vLLM's \texttt{FP8} kernels) and at least 16~GB of memory: the served MINE models occupy 3.5--10.5~GB, tuning processes one launch at a time, and quantization needs up to about 20~GB (24~GB recommended). Plain vLLM on the \texttt{BF16} model, ZipServ's vLLM~0.9.1 on the \texttt{BF16} model, and the vLLM arm of the prefill experiment need the uncompressed weights on the GPU (24~GB or more); on smaller GPUs they are not run, as for the RTX~4060~Ti in Figure~\ref{fig:throughput}. ZipServ fits in 16~GB with the near-capacity settings of the MINE serves, but its batch-1 check then uses the dense \texttt{BF16} tokens of a run on a larger GPU as reference. For the decode runs, the vLLM~0.27.1 KV cache is fixed at 2~GiB on GPUs with less than 24~GB; otherwise, vLLM sizes it at 85\% utilization. The prefill runs fix it at 1~GiB on every GPU. The setup builds vLLM for MINE from source for the GPUs present, as for the paper; the baselines use the official vLLM~0.27.1 wheel, which covers sm\_75 to sm\_120. The tuner's tile menus were chosen on our two GPUs; on others it still searches all legal configurations. Because tune files contain measured timings, a tune file must only be served on the GPU model it was tuned on. The complete pipeline needs about 300~GB of disk, 64~GB of host memory, CUDA toolkits 13.x and 12.8/12.9 (the latter only to build NeuZip and ZipServ), and takes 3--4~hours of setup, about 7~hours for the checkpoints and statistics, and 1.5--2~days per GPU.

\paragraph{Modifications of third-party code.} We changed third-party code only where it did not build or run in our setting: one \texttt{constexpr} removed in NeuZip (nvcc~12 otherwise compiles its Python bindings for the device); ZipServ's compressor built for the local architecture instead of the hard-coded sm\_89; in ZipServ's vLLM integration, its GEMM kernel sources replaced by those of the ZipServ artifact, because the integrated copy skips the computation for fewer than 8~tokens and thus at batch size 1, and in that kernel two function-local \texttt{static} shared-memory sizes computed per launch instead of once (the first layer launched fixed them; with the \texttt{lm\_head} compressed, later and larger layers overran shared memory), plus a build-only fix to locate NVTX3; auto-round's calibration set pinned to a dataset revision by a wrapper. Three runtime wrappers only change where memory is used, never a result: EntQuant's reported entropy statistic is computed with a histogram on the host (during quantization and when its nvCOMP baseline loads), its compressed buffers are moved to the host during quantization, and GPTQModel's CPU fallback after an out-of-memory error in its channel permutation, which would permute the weights twice, is refused.

\end{document}